\pdfoutput=1

\documentclass[11pt]{article}

\usepackage[]{ACL2023}

\usepackage{times}
\usepackage{latexsym}
\usepackage{amssymb}
\usepackage{pgfplots}
\usepackage{scalefnt}
\usepackage{boldline}
\usepackage{xcolor}
\usepackage{multirow}
\usepackage{makecell}
\usepackage{graphicx}
\usepackage{bm}
\usepackage{amsmath}
\usepackage{bbm}
\usepackage{float}
\usepackage{stfloats}
\usepackage{lipsum}
\usepackage{comment}
\usepackage{booktabs}
\usepackage{subfig}
\usepackage{lipsum}
\usepackage[flushleft]{threeparttable}

\usepackage{tabularx}
\usepackage[T1]{fontenc}

\usepackage[utf8]{inputenc}

\usepackage{microtype}

\usepackage{inconsolata}

\newcommand\blfootnote[1]{%
  \begingroup
  \renewcommand\thefootnote{}\footnote{#1}%
  \addtocounter{footnote}{-1}%
  \endgroup
}

\newcommand{\cdiff}{ClaimDiff}
\newcommand{\cdiffs}{ClaimDiff-S}
\newcommand{\cdiffw}{ClaimDiff-W}

\definecolor{purple1}{HTML}{7570b3}
\definecolor{green1}{HTML}{1b9e77}
\definecolor{orange}{HTML}{d95f02}
\definecolor{green2}{HTML}{4daf4a}
\definecolor{red}{HTML}{e41a1c}
\definecolor{blue}{HTML}{377eb8}
\definecolor{purple2}{HTML}{984ea3}

\newcommand{\draftonly}[1]{#1}
\renewcommand{\draftonly}[1]{}

\newcommand{\draftcomment}[3]{\draftonly{\textcolor{#2}{{\textbf{[#3 --\textsc{#1}]}}}}}

\newcommand{\miyoung}[1]{\draftcomment{miyoung}{cyan}{#1}}

\title{ClaimDiff: Comparing and Contrasting Claims on Contentious Issues}

\author{Miyoung Ko$^{a, \star}$ \quad  Ingyu Seong$^{b, \dagger}$ \quad Hwaran Lee$^c$ \quad \\
        \textbf{Joonsuk Park}$^{c, d}$ \quad \textbf{Minsuk Chang}$^{c, \ddagger}$ \quad \textbf{Minjoon Seo}$^{a}$\\ 
        $^a$KAIST \quad $^b$Korea University \quad $^c$ NAVER AI Lab 
        \quad $^d$ University of Richmond \\
\texttt{\{miyoungko, minjoon\}@kaist.ac.kr} \quad \\
\texttt{dlssrb7777@korea.ac.kr} \quad 
\texttt{park@joonsuk.org} \quad\\
\texttt{\{hwaran.lee, minsuk.chang\}@navercorp.com} 
}

\begin{document}
\maketitle
\begin{abstract}
    With the growing importance of detecting misinformation, many studies have focused on verifying factual claims by retrieving evidence. 
However, canonical fact verification tasks do not apply to catching subtle differences in factually consistent claims, which might still bias the readers, especially on contentious political or economic issues.
Our underlying assumption is that among the trusted sources, one's argument is not necessarily more true than the other, requiring \emph{comparison} rather than \emph{verification}.
In this study, we propose \cdiff, a novel dataset that primarily focuses on \emph{comparing} the nuance between claim pairs. 
In \cdiff, we provide 2,941 annotated claim pairs from 268 news articles. 
We observe that while humans are capable of detecting the nuances between claims, strong baselines struggle to detect them, showing over a 19\% absolute gap with the humans.
We hope this initial study could help readers to gain an unbiased grasp of contentious issues through machine-aided comparison.

\end{abstract}

\blfootnote{\textsuperscript{$\star$}This work was done during internship in NAVER AI Lab.}
\blfootnote{\textsuperscript{$\dagger$}Is now at Samsung Research.}
\blfootnote{\textsuperscript{$\ddagger$}Is now at Google.}

\section{Introduction}
\label{sec:intro}

With an ever-increasing amount of textual information on the web,
many researchers have focused on detecting misinformation from diverse sources, such as fake news \cite{potthast-etal-2018-stylometric, 10.1145/3340531.3412046} and rumor tweets \cite{zubiaga2016analysing, kochkina-etal-2018-one}.
In particular, fact verification has become a popular task due to its utility and the availability of reliable datasets such as FEVER \cite{thorne-etal-2018-fever, aly2021feverous}. 

\begin{figure*} [t]
    \centering
    \subfloat
    {\includegraphics[width=0.74\textwidth]{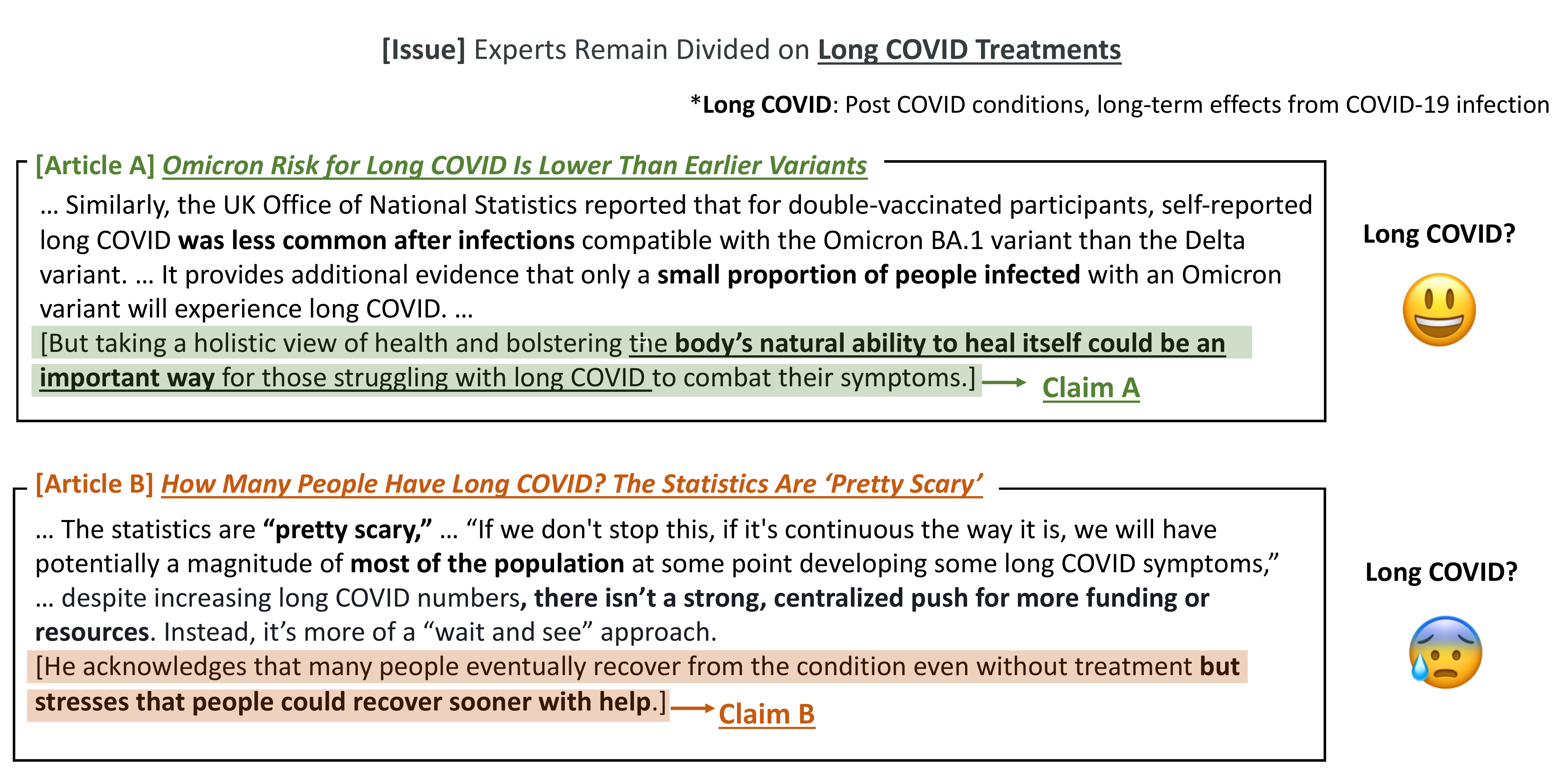}}\hfill
    \subfloat
    {\includegraphics[width=0.26\textwidth]{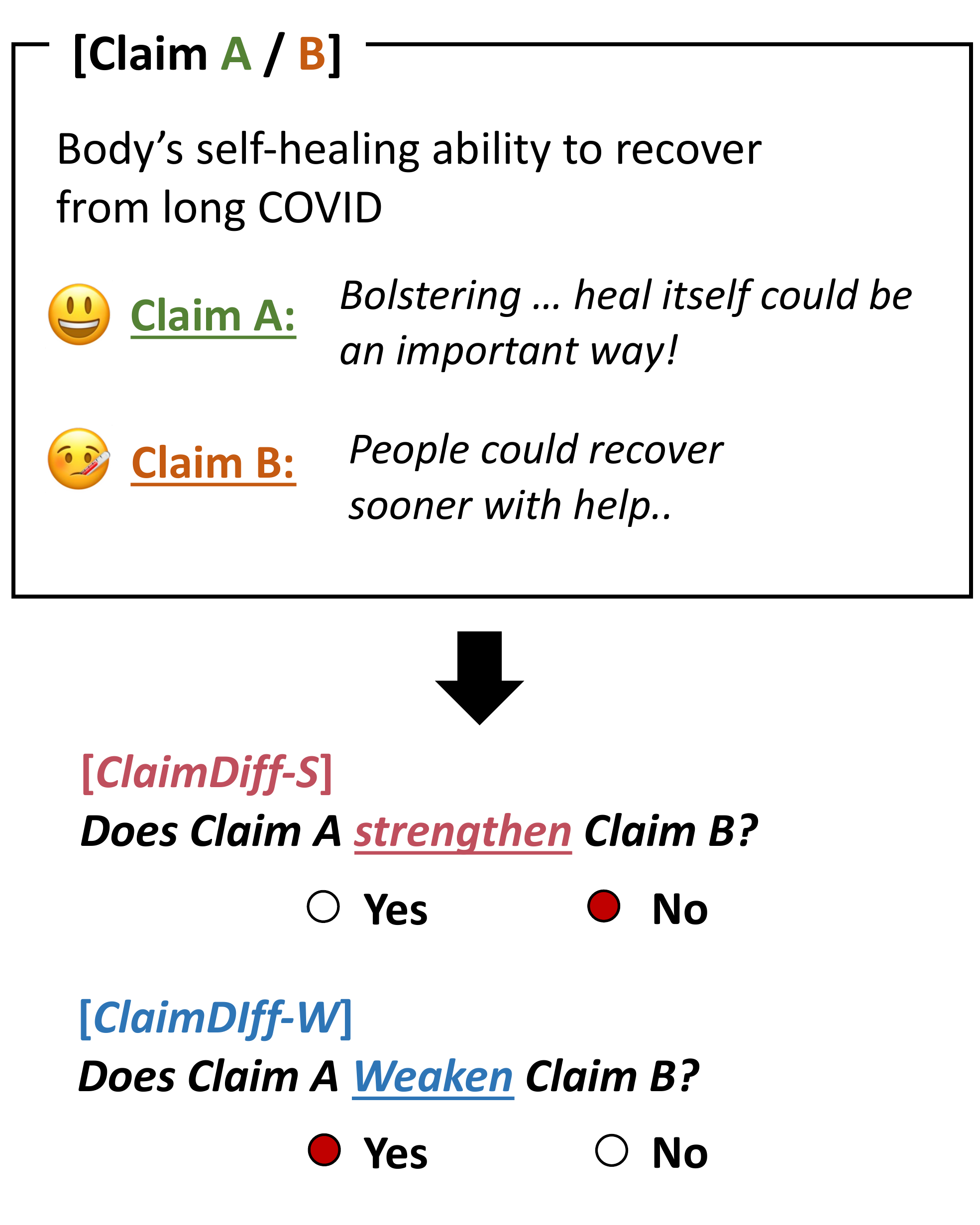}}
     \vspace{-0.2cm}
    \caption{Two articles are reporting on long COVID with opposite perspectives: (A) Recent variants have less risk for long COVID, while (B) Statistics for long COVID is \textit{pretty scary}. This might lead readers to have different understandings of long COVID. \cdiff~targets the "strengthening" and "weakening" relationships between two claims on the same issue. 
    Although both claims are about self-healing ability to overcome long COVID, the nuances are different, having "weakening" relation.
    The rationale for 'A weakens B' relation is underlined in Claim A. 
    }\vspace{-0.4cm}
    \label{fig:fig1_teaser}
\end{figure*}

For contentious issues, however, fact verification alone is not sufficient; comparing and contrasting claims from the opposing sides are necessary to gain an unbiased understanding of the issue.
Figure~\ref{fig:fig1_teaser} presents two articles reporting on the treatment of long COVID -- long-term physical and mental symptoms that can occur after COVID-19 infection.
Although both articles are published in the same week, their perspectives on long COVID are quite different; article A downplays the risk of long COVID stating that it has become less common in the recent COVID variants, while article B expresses concerns about the increase in the number of people suffering from long COVID.
In this way, even articles from trusted sources may provide biased views about an issue. 

In this paper, we present \cdiff, a novel dataset consisting of 2,941 claim pairs extracted from 268 news articles on 134 contentious issues.\footnote{The articles are collected from allsides.com, licensed under a CC BY-NC 4.0 license.}
{\cdiff} comes in two variations---{\cdiffs} and {\cdiffw}---each consisting of labels targeting a different relation: determining whether a claim \textbf{S}trengthens and \textbf{W}eakens another claim, respectively.
For instance, the two claims in Figure~\ref{fig:fig1_teaser} weaken each other, providing inconsistent perspectives surrounding the undisputed factual information --- the human body has a natural ability to heal from long COVID. 
More specifically, claim A argues that medical treatments should be geared toward supporting the natural ability to heal, while claim B calls for more active interventions for fast recovery.
\cdiff~focuses on recognizing such relations between two claims on an issue; this is distinguished from existing tasks such as fact verification --- verifying the veracity of a claim using evidence text~\cite{ vlachos-riedel-2014-fact, thorne-etal-2018-fever} --- and stance detection --- identifying the stance of a claim toward a topic of interest~\cite{mohammad-etal-2016-semeval, derczynski-etal-2017-semeval}.

We also demonstrate the efficacy of {\cdiff} on two tasks it supports --- relation classification and rationale extraction --- and an extended application --- document-level \cdiff.
Relation classification experiments tests the ability to identify strengthen and weaken relations.
Rationale extraction tests the ability to find the rationale for a given relation in extractive and generative scenarios.
Lastly, although \cdiff~only provides sentence-level annotations, models trained on it allow document-level analyses.
In this document-level \cdiff, 
we test the potential of analyzing points of agreement and disagreement in documents.

The main contributions of our work are:
\begin{itemize}

    \item We present \cdiff, a novel dataset of claim pairs on contentious issues. Each pair comes with strengthen and weaken relation labels, as well as a rationale for choosing the label.\footnote{\url{https://github.com/kaistAI/ClaimDiff}}
    \item We present competitive baselines for the dataset, leveraging finetuning, parameter-efficient finetuning, and zero-shot approaches, along with human performance. 
    \item We showcase how models trained on \cdiff~can be used to compare documents on an issue, indicating specific points of agreement and dispute, in document-level \cdiff. 

\end{itemize}

\section{Related Works}
\label{sec:related_work}

\paragraph{Dealing with Misinformation}
Detecting and avoiding misinformation received intense interest with a massive amount of information on the web.
Existing works introduced benchmarks with a broad spectrum of sources, including rumors in social media \cite{potthast-etal-2018-stylometric, kochkina-etal-2018-one} and fake news \cite{zubiaga2016analysing}.
Other researchers focus on dealing with an exploding amount of misinformation on global events, such as the COVID-19 pandemic \cite{saakyan-etal-2021-covid, jiang2021categorising, alam-etal-2021-fighting-covid, 10.1145/3485447.3512039}. 
While many existing benchmarks aim to detect less reliable information on unverified sources, our work targets subtle differences in trusted sources.

\paragraph{Claim Verification}
Claim verification verifies the factuality of the target sentence with respect to a reliable truth from verified sources. 
Automatic verification shows remarkable progress with the introduction of rich claim verification datasets \cite{vlachos-riedel-2014-fact, thorne-etal-2018-fever, hanselowski-etal-2019-richly, aly2021feverous, khan-etal-2022-watclaimcheck}. 
Existing claim verification datasets introduce many variants, including a shift in domains and languages of claims; claims from political sources \cite{wang-2017-liar, 10.1145/3178876.3186139}, scientific claims \cite{wadden-etal-2020-fact}, climate change-related claims \cite{leippold2020climatefever}, Arabic claims \cite{baly-etal-2018-integrating, alhindi-etal-2021-arastance}, and Danish claims \cite{norregaard-derczynski-2021-danfever}.
Our assumption is different from claim verification in that one claim is not necessarily more true than the other.
As a result, \cdiff~focues on comparison between claims rather than verification.

\paragraph{Stance Detection}
Stance detection aims to predict the stance of a claim toward a specific topic between agreeing or opposing perspectives.
\citet{mohammad-etal-2016-semeval} propose SemEval challenges to predict the stance of tweets toward target keywords. 
\citet{derczynski-etal-2017-semeval} further presents a sub-challenge to detect the stance of corresponding threads of rumor tweets.
Other benchmarks are introduced with diverse challenges, such as claim-based stance detection \cite{ferreira-vlachos-2016-emergent, bar-haim-etal-2017-stance}, stance detection with evidence \cite{chen-etal-2019-seeing}, and stance detection over political domains \cite{li-etal-2021-p}.

\section{\cdiff}
\label{sec:method}

In this section, we formally define the tasks supported by \cdiff, 
describes how the dataset is constructed,
and provide the statistics and analysis of the resulting dataset.

\subsection{Task Description}\label{sec3:task}

{\cdiff} comes in two variations---{\cdiffs} targeting strengthen relations, and {\cdiffw} targeting weaken relations. Both variations of the dataset were designed to support the following tasks.



\paragraph{Relation Classification}
Relation classification aims to determine if claims from two different documents are in a relation: strengthen for \cdiffs, and weaken for \cdiffw.
For instance, as shown in Figure~\ref{fig:fig1_teaser}, claim A and claim B are both about the treatment of long COVID. 
Claim A and B exhibit opposing positions for the body's natural recovery, respectively.
We want to classify this case into \textit{weakens} as claim A weakens claim B. 
Although, in this case, claim B weakens A as well, note that the relationship is not guaranteed to be symmetric in general. 

More formally, for \cdiffs, given a claim pair $(c_1, c_2)$, the objective is to return \textit{true} if $c_1$ strengthens $c_2$, and \textit{false}, otherwise. 
For \cdiffw, given a claim pair $(c_1, c_2)$, the objective is to return \textit{true} if $c_1$ strengthens $c_2$, and \textit{false}, otherwise.
Note that for any given claim pair, the answer cannot be \textit{true} for both variations of \cdiff.

\paragraph{Rationale Extraction}
Rationale extraction aims to extract phrases from a claim in a relation: strengthen for \cdiffs, and weaken for \cdiffw.
For instance, in Figure~\ref{fig:fig1_teaser}, a rationale for \cdiffw~is \textit{`body’s natural ability to heal itself could be an important'},
which weakens claim B about the view that more active interventions for fast recovery.

More formally, given a claim pair $(c_1, c_2)$ in a relation, the objective is to extract phrases in $c_1$ that provide a rationale for relation.
Note that a single pair of claims can have multiple rationales.

\subsection{Constructing \cdiff~Dataset}\label{sec3:construction}

\paragraph{Raw Data Collection}
\label{sec:raw_data}
We first collect a group of news articles from AllSides\footnote{\url{https://www.allsides.com/unbiased-balanced-news/}} headlines.
Allsides provides a balanced search of news with all sides of the political spectrum. 
More specifically, in AllSide headlines, there are groups of news articles about the same issues from different media press.
The article groups have a broad political spectrum; each belongs to one of (left, center, and right) political stances.
We crawl the headline pages uploaded from 2012-06-01 to 2021-11-21, covering more than 180 media sources.
We choose two articles with left and right labels if possible.

To construct the claim pairs, we filter the non-overlapping sentence pairs from the article pair.
As most claim pairs do not provide overlapping contents, a large proportion of pairs are filtered out. 
We apply an additional filtering process using Amazon Mechanical Turk (MTurk) \footnote{\url{https://www.mturk.com/}} to collect the overlapping claim pairs. 
Each worker is asked to answer the question, \textit{'Does the target sentence overlap with a given sentence?'}, where each sentence is extracted from two different articles. 
Given a single example, three workers made a response. 
We collected the pairs if at least two workers answered that the given pairs are "overlapping". 
The details of the filtering process are shown in Appendix~\ref{appendix:data_collect}.

\paragraph{Annotation Process}
After the filtering process, we conducted an annotation process with 15 in-house expert annotators to obtain the final data. 
Given a pair of claims, the annotators were requested to determine the stance
among \textit{strengthen}, \textit{weaken} and \textit{no effect}. 
If they choose \textit{strengthen} or \textit{weaken}, the annotators had to select the phrases from the claim that strengthen or weaken the other claim. 
The overall interface for the data collection process is shown in Figure~\ref{fig:appendix_annot}.

For each single claim pair, three to five participants submitted their responses. 
We collect the responses and convert the relation options to scaled values. 
We first consider \textit{strengthen} as 1, \textit{weaken} as -1, and \textit{no effect} as 0 and average the choices after conversion.
We filter out the pairs with absolute average values between (0, 0.5), which means the relations are ambiguous.\footnote{Unfiltered data with average response scores are also publicly available.}
The pairs with positive values are mapped into strengthening claims (1 for \cdiffs), and the pairs with negative values are mapped into weakening claims (1 for \cdiffw).  
If the resulting values are exactly 0, the claims become 0 for both \cdiffs~and \cdiffw.  
We measure the average inner-annotation agreement by Krippendorff’s alpha \cite{doi:10.1080/19312450709336664}.
The scores are 0.46 and 0.47 for \cdiffs~and \cdiffw, respectively.

\begin{table} [t]
\centering
\resizebox{0.4\textwidth}{!}{
\def\arraystretch{0.9}
\begin{tabular}{l|c|c|c}
\toprule
& \textbf{Train} & \textbf{Test}&\textbf{Test-doc}\\\midrule
Pairs & 1,857 & 1,084 & 3,173 \\
Issues & 90 & 44 & 44 \\ 
Articles & 180 & 88 & 88 \\ 
Rationales & 1,484 & 852 & 852 \\
\midrule
\% Strengthen & 69.31\% & 56.64\% & 19.35\% \\
\% Weaken & 10.61\% & 21.96\% & 7.50\%\\
\bottomrule
\end{tabular}}
\vspace{-0.05cm}
\caption{Statistics of \cdiff\ dataset. 
	Test-doc indicates the raw test dataset over the whole article, including non-overlapping claims.}\vspace{-0.4cm}
\label{table:stat}
\end{table}

\paragraph{Constructing Test-doc Dataset}
We aim to build an application for understanding contentious issues with diverse views on a fine-grained level. 
Rather than classifying an article in a single label (i.e., containing left or right political bias), \cdiff~enables a comparison between two articles in a sentence-wise manner. 
However, this requires a further extension of \cdiff~from sentence pairs to document pairs, having significantly different distributions. 
To be coherent with the real-world distribution over whole articles, we provide an additional test dataset, \textit{test-doc}, following the distribution over article pairs.
Test-doc can be considered as an unfiltered test dataset, resulting in a high ratio of non-overlapping pairs.  
We collect the non-overlapping claim pairs of articles in the test dataset that are obtained from the previous filtering step.
Over 70\% of the claim pairs from article pairs are not overlapped, resulting in a highly skewed distribution. 
We combine filtered-out claim pairs with a standard test dataset to construct the final test-doc dataset.

\subsection{Dataset Statistics}\label{sec3:stat}

We extract the pairs from a group of articles sharing the same topic, published by multiple presses. 
Our final dataset contains articles from 47 presses.
The top-3 presses with the highest appearance are \textit{Fox News}, \textit{CNN}, and \textit{Washington Times}. 
We further analyze the topic diversity of our dataset by collecting the tag information. 
\textit{Tag} is the list of words representing the topic provided in Allsides headlines. 
For instance, tags for a topic, \textit{Supreme Court Sides With Google in Copyright Dispute Case}, are \textit{Supreme Court}, \textit{Copyright}, \textit{Google}, and \textit{Oracle}. 
The number of unique tags is 276, while each topic includes an average of 3.6 tags.
The most common issues are \textit{Elections}, \textit{Donald Trump}, and \textit{Coronavirus}.
The overall lists of presses and tags are presented in Appendix~\ref{appendix:example}. 

Table~\ref{table:stat} presents the overall statistics of \cdiff. 
\cdiff~dataset provides 2,941 examples, extracted from 268 articles with 134 issues. 
Note that \cdiffs~and \cdiffw~consists of the same claim pairs with different labels. 
Since non-overlapping claim pairs do not provide rationales, the number of rationales is less than the overall claim pairs. 
Each pair contains an average of 1.4 rationales with an average length of 13.2 tokens.\footnote{We use spaCy tokenizer for tokenizing the rationales.}
In the train and test dataset, "strengthening" pairs are available to be found with more than 50\% appearance. 
Finding the "weakening" claim pairs is more challenging, resulting in 7.50\% weakening examples in the test-doc environment.

\subsection{Dataset Analysis}\label{sec3:analysis}

This part analyzes claim pairs in \cdiff~with respect to the class label.
We first provide the subjectivity analysis over claims with positive labels in \cdiffs~and \cdiffw. 
\cdiff~includes both subjective and objective claims, indicating the proposed task is designed to predict a  more general relation between each claim pair.
We further present prediction results of the natural language inference (NLI) and fact verification (FEVER) models.
The results indicate that models trained on the datasets are not suitable for understanding the nuances.\footnote{Check the Appendix~\ref{appendix:data_anal} for additional information.}

\paragraph{Subjectivity Analysis}
To analyze the subjectivity of claims, we randomly sample 50 "strengthening" examples from \cdiffs~and 50 "weakening" examples from \cdiffw~test data. 
We manually label each claim in a pair with \textit{subjective} or \textit{objective}. 
Following \citet{10.1007/978-3-540-30586-6_53}, we distinguished subjective and objective claims based on whether each claim includes at least one private state -- opinions, evaluations, emotions, and speculations.
We found that "strengthening" pairs have a high proportion of objective claims on both claim A and claim B (A: 72\%, B: 80\%). 
However, "weakening" examples include more than 40\% of subjective claims (A: 44\%, B: 48\%), indicating more diverse patterns in "weakening" relations.  
We expect that \cdiffw~to be more challenging not only because of the skewed distribution but also because of the more diverse composition of claims. 

\paragraph{Prediction Results of NLI / FEVER Model}

To compare the \cdiff~with previous sentence pair classification tasks, we analyze prediction results of transformer-based models trained on NLI and FEVER. 
We use RoBERTa-large trained on MNLI \cite{williams-etal-2018-broad} and FEVER \cite{thorne-etal-2018-fever}.
Each model yielded 90.2 (MNLI) and  75.6 (FEVER) F1 scores, respectively. 
Among  614 "strengthening" and 239 "weakening" pairs, the MNLI model predicts 561 and 203 pairs as \textit{neutral}. 
FEVER model predicts 532 (strengthen) and 226 (weaken) examples as \textit{not enough info}.
These failures might come from multiple reasons, including the domain shift in claims and the different goals of each task. 
However, we observe that "weakening" pairs contain a slightly higher ratio of \textit{contradiction} (strengthen: 5.7\% vs. weaken: 15\%) and 0 \textit{entailment} with the MNLI model. 
"Strengthening" pairs show the difference in FEVER, containing more \textit{support} (FEVER, 9.0\% vs. 1.6\%) examples.
MNLI and FEVER models might be able to distinguish weakening and strengthening examples from others, which can work like the prior knowledge for solving \cdiff~ in Section~\ref{sec:experiments_class}.

\section{Task 1 - Relation Classification}
\label{sec:experiments_class}

\subsection{Experimental Setup}\label{sec:sw_setup}

In this section, We present the baselines with different learning strategies: (1) finetuning, (2) parameter-efficient finetuning, and (3) zero-shot baselines.
Implementation details of each model are described in Appendix~\ref{appendix:imp_detail}. 

\paragraph{Finetuning Baselines}
We finetune pre-trained language models for the sentence classification tasks. 
Each model takes a pair of claims as inputs and predicts whether the first claim strengthens/weakens the other. 
We finetune models with a weighted loss function, where the class weight is determined by the label distribution.\footnote{Training without the weight is not possible due to the extremely skewed distribution of \cdiffw, resulting 0 F1.}
We construct the development data by sampling 30\% of issues from training data. 
Following the test environment, issues in development data are exclusive.
We train RoBERTa-base and RoBERTa-large \cite{liu2019roberta} on \cdiffs~and \cdiffw, respectively. 
We further present RoBERTa (FEVER) and RoBERTa (MNLI), RoBERTa-large initialized on MNLI and FEVER.\footnote{We use the same models as in Section~\ref{sec3:analysis} for initialization.}

\begin{figure} [t]
    \centering\vspace{-0.6cm}
    \subfloat
    {\includegraphics[width=0.41\textwidth]
    {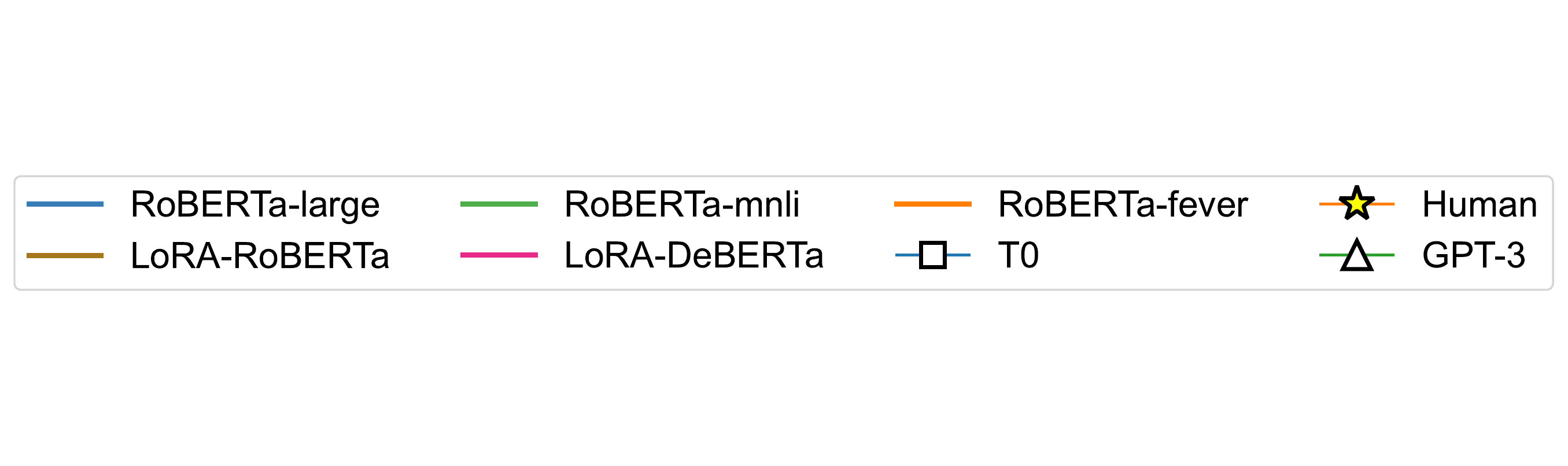}} \\[-3ex]
     \subfloat[\cdiffs]
     {\includegraphics[width=0.41\textwidth]{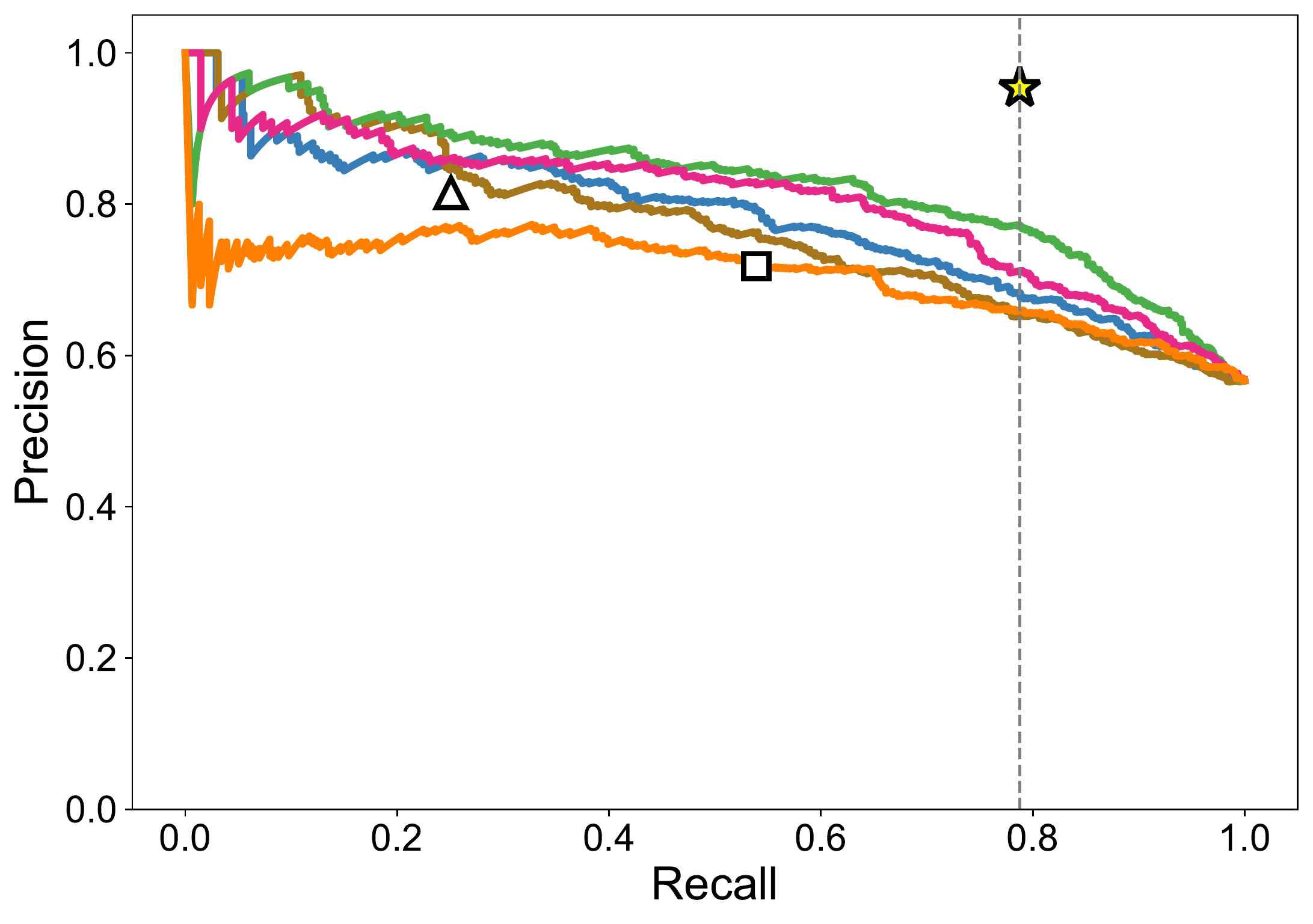}}\\\vspace{-0.1cm}
     \subfloat[\cdiffw]
     {\includegraphics[width=0.41\textwidth]{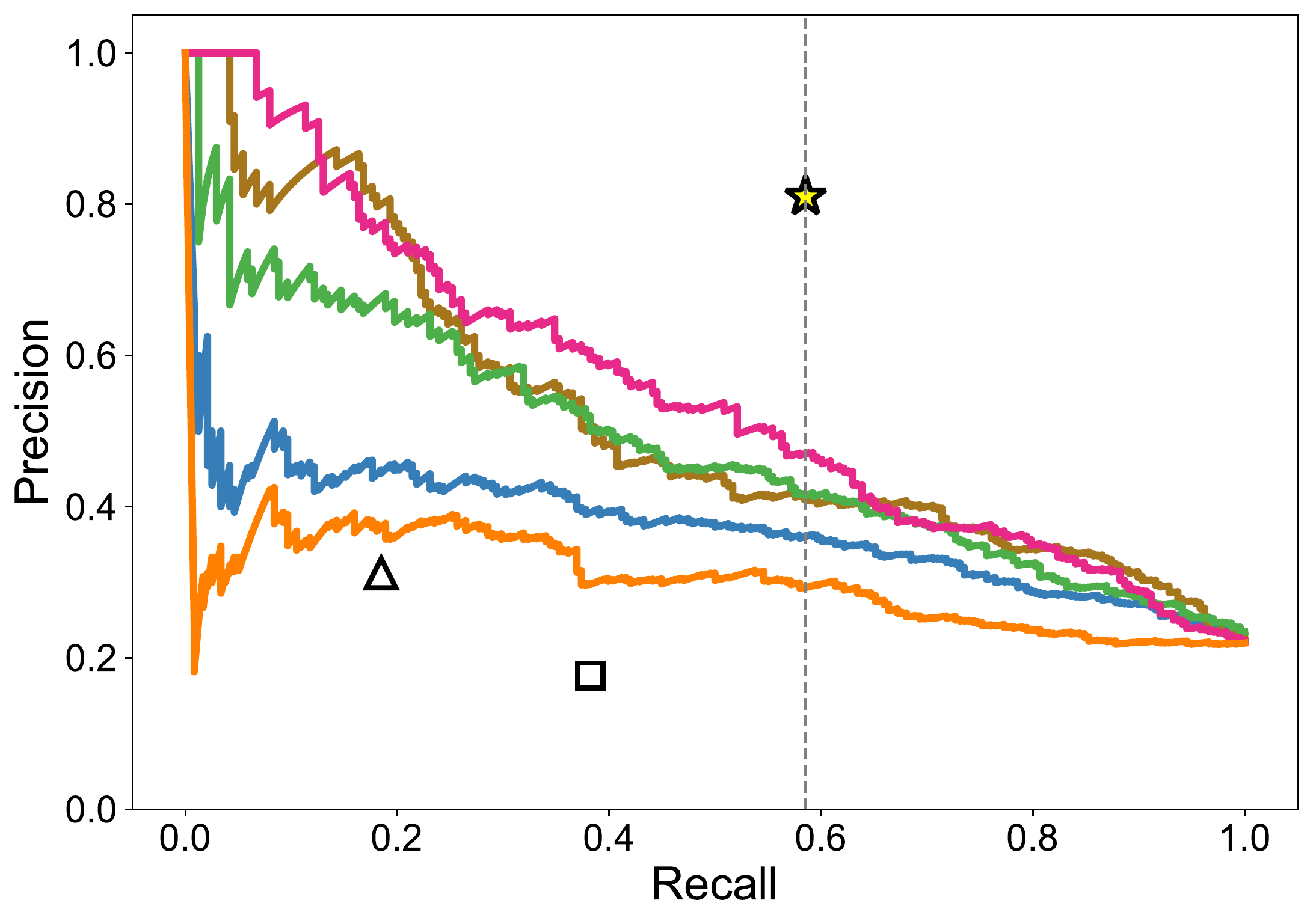}}
     \vspace{-0.1cm}
    \caption{Precision-Recall curve of the baselines.
    Zero-shot baselines are represented as points as it is not feasible to control threshold for zero-shot predictions.
    }\vspace{-0.5cm}

    \label{fig:pr_curve}
\end{figure}

\paragraph{Parameter-efficient finetuning}
As the number of training examples is not large enough, we further explore the parameter-efficient finetuning methods. 
Following \citet{hu2022lora}, we apply low-rank adaptation (LoRA) on pre-trained language models, which only finetune the task-specific low-rank matrices. 
We follow the same procedure as finetuning baselines for model selection. 
We compare the effect of LoRA on RoBERTa-large (355M) and DeBERTa-XXL (1.5B) \cite{he2021deberta}. 

\begin{table*} [t]
\centering\
\resizebox{0.9\textwidth}{!}{
\def\arraystretch{0.95}
\begin{tabular}{l|c|cccc|cccc}
\toprule
\multirow{2}{*}{Model}
& \# Trainable  & \multicolumn{4}{c|}{\textit{\cdiffs}} & \multicolumn{4}{c}{\textit{\cdiffw}} \\
& Param. & AUROC & F1 & Precision & Recall 
& AUROC & F1 & Precision & Recall \\\midrule
\textbf{\textit{finetuning}} & & & & & & & & & \\
~RoBERTa-base & 125M /125M & 0.7160 & 71.82 & 68.97 & 74.92 & 0.6350 & 38.04 & 37.05 & 39.08\\
~RoBERTa-large & 355M / 355M & 0.7085 & 71.37 & 67.49 & 75.73 & 0.7137 &44.89 & 35.53 & 60.92 \\
~RoBERTa (FEVER) & 355M /355M & 0.6841 & 72.61 & 64.21 & 83.55 & 0.6056 & 36.77 & 24.29 & \textbf{75.63} \\
~RoBERTa (MNLI) & 355M / 355M & \textbf{0.7976} & \textbf{77.86} & 70.80 & \textbf{86.48} & 0.7481 & 45.33 & 48.11 & 42.86 \\
\midrule
\textbf{\textit{LoRA}} & & & & & & & & & \\
~RoBERTa-large & 0.8M / 355M & 0.5194 & 61.83 & 55.34 & 70.03 & 0.7633 & 42.13 & 49.71 & 36.55 \\
~DeBERTa-XXL & 4.7M / 1.5B & 0.7668 & 76.34 & 69.72 & 84.36 & \textbf{0.7685} & \textbf{49.07} & \textbf{54.64} & 44.54\\
\midrule
\textbf{\textit{zero-shot}} & & & & & & & & &  \\
~T0 & 0 / 11B & -$^*$ & 61.58 & 71.80 & 53.91 & -$^*$  & 24.17 & 17.67 & 38.24\\
~GPT-3 & 0 / 175B & -$^*$  & 38.36 & \textbf{81.48} & 25.08 & -$^*$  & 23.22 & 31.21 & 18.49\\\midrule
\textbf{\textit{Human Evaluation}} & - & - & 86.27 & 95.34 & 78.77 & - & 68.29 & 81.01 & 58.56\\
\bottomrule
\end{tabular}}
\vspace{-0.05cm}
\caption{Performance on test dataset. \textit{\# Trainable Param.} represents the number of trainable parameters over the number of model parameters. 
Human indicates the human performance evaluated on test dataset. (*AUROC is not defined as threshold is not applicable for zero-shot generation.)
}\vspace{-0.35cm}
\label{table:test}
\end{table*}

\paragraph{Zero-shot Baselinse}
We present the zero-shot performance of large language models, T0 \cite{wei2021finetuned} and GPT-3 \cite{NEURIPS2020_1457c0d6, Ouyang2022TrainingLM}.
Both models get a test pair following the prompt format and generate the answer directly. 
We consider the class token with maximum probability as the prediction results. 
The examples of the zero-shot prompts are presented in Appendix~\ref{appendix:imp_detail}.

\subsection{Evaluation on \cdiff~Test}

\paragraph{Human Evaluation}
Following \citet{rajpurkar-etal-2016-squad}, we evaluate human performance on \cdiff~based on the human annotation results. 
As each example has at least three responses, we randomly sample one response as the human prediction.
We obtain ground-truth labels using the remainder following the same procedures as in Section~\ref{sec3:construction}.
The resulting human performance is shown in Table~\ref{table:test}. 
Humans are capable of detecting both types of relation, resulting in a significant gap between human and model performance. 
Even for more challenging \cdiffw, humans can detect more than half of the weakening nuances while maintaining 81.01\% precision.

\paragraph{Main Results}

Figure~\ref{fig:pr_curve} shows the precision-recall curve of our baselines. 
Compared to the vanilla RoBERTa-large, initialization with FEVER worsen the performance while MNLI gives a significant gain in both \cdiffs~and \cdiffw.
Parameter-efficient finetuning (LoRA) is effective for \cdiffw, even when using a same size model (RoBERTa-large). 
This is due to the small number of "weakening" examples, which makes finetuning the whole parameters more difficult. 
When recovering about 80\% of "strengthening" examples (\cdiffs), humans retain over 90\% of precision, while the best working model retains 80\%. 
The gap is more significant in \cdiffw, showing about a 30\% of difference in precision.
Zero-shot baselines, T0 and GPT-3, are worse than finetuning RoBERTa-large in both tasks.
We use prompts asking about a single relation ("support" / "weaken") for zero-shot baselines, while the actual relations in \cdiff~are more complex, which results low coverage of zero-shot models (i.e., GPT-3 predicts only 17\% of examples as "strengthening").

The overall evaluation results are presented in Table~\ref{table:test}. 
Note that we report AUROC except for zero-shot baselines, as zero-shot generation does not require any threshold. 
Align with previous observations, initialization by MNLI boost the performance on both tasks, obtaining the best AUROC in \cdiffs.
LoRA becomes more effective when the number of examples is small.
LoRA enables training the 1.5B size model (DeBERTa-XXL) with only the hundreds of "weakening" examples, obtaining the best AUROC and F1 in \cdiffw.

\paragraph{Error Analysis}
To further understand the challenges in \cdiff, we investigate the errors of finetuned RoBERTa-large. 
We randomly sample 25 false negatives (i.e., relationships that the model failed to detect) from each \cdiffs~and \cdiffw.\footnote{We also analyzed the false positives (FP) while the patterns of FP are too diverse to capture the common patterns.}
We manually categorize the errors and analyze them.
The examples of each category and the ratio are provided in Appendix~\ref{appendix:error_analysis}.
The errors in "strengthening" have relatively simple patterns (ex., 28\% entailment or 20\% coherent nuance).
However, it is more difficult to capture these relationships in \cdiff~as claims are collected from real-world news articles, resulting relatively low lexical overlap between two claims. 
On the other hand, we find that patterns in \cdiffw~are more challenging; we suspect that this is because there are diverse ways to weaken one's argument. 
Finally, a common error (12\%) in \cdiffs~and \cdiffw~is due to the need for context information or background knowledge to understand the claims.


\section{Task 2 - Rationale Extraction}
\label{sec:experiments_rationale}

We perform rationale extraction over "strengthening" and "weakening" pairs, which have positive labels for \cdiffs~and \cdiffw. 
Because of the low appearance of "weakening" pairs, training individual models for each task is challenging. 
Therefore, unlike relation classification, we train a single model to extract rationales from both "strengthening" and "weakening" pairs.
We experiment with extractive and generative models. 
The evaluation is also conducted on combined sets of \cdiffs~and \cdiffw.

\subsection{Models}
\paragraph{Extractive Model} 
In this work, we present machine reading comprehension (MRC) models as the extractive baselines.
As shown in Figure~\ref{fig:fig1_teaser}, rationales are found as the phrases existing in input claims. 
Extracting the phrases from a given text is similar to the previous MRC \cite{rajpurkar-etal-2016-squad, trischler-etal-2017-newsqa}. 
Following \citet{devlin-etal-2019-bert}, we finetune pre-trained language models with the output layer that predicts the start and end positions of given rationales. 
We train RoBERTa-base and RoBERTa-large for rationale extraction.

\paragraph{Generative Model}
Generative baselines directly \textit{generate} the rationales rather than extract it from input claims. 
Existing works \cite{narang2020wt5, lakhotia-etal-2021-fid} show that generative models obtain a strong performance on rationale benchmark, ERASER \citep{deyoung-etal-2020-eraser}. 
ERASER is designed to evaluate the reasoning ability of NLP models,  containing 7 NLP tasks, including BoolQ \cite{clark-etal-2019-boolq}, and Movie Reviews \cite{zaidan-eisner-2008-modeling}. 
Following \citet{narang2020wt5}, we finetune the T5 models to sequentially generate a list of rationales in a token-by-token fashion. 
We experiment with T5 and T5 with "strengthening" / "weakening" labels. 
The exact input formats are explained in Appendix~\ref{appendix:imp_detail}.

\begin{table} [t]
\centering\
\resizebox{0.4\textwidth}{!}{
\def\arraystretch{0.9}
\begin{tabular}{l|ccc}
\toprule
& \textbf{Perplexity} & \textbf{TF1} & \textbf{IOU F1}\\\midrule
RoBERTa-base & -  & 63.67 & 57.25\\
RoBERTa-large & -  & 63.49 & 55.99\\
T5-base & 1.49 & 72.78 & 65.05 \\
~ + \textit{class label} & 1.45 & 72.70 & 63.95 \\
T5-large & 1.45 & 75.08 & \textbf{67.74}\\
~ + \textit{class label} & 1.46 & \textbf{77.01} & \textbf{67.70}\\ 
\bottomrule
\end{tabular}}
\vspace{-0.05cm}
\caption{Rationale extraction performance measured on test set.
Note that it is possible to measure perplexity only for generative baselines (T5). 
 }\vspace{-0.35cm}
\label{table:rationale}
\end{table}

\subsection{Evaluation Metrics}

\paragraph{Perplexity}
We report the per-token perplexity of rationales that measures how well the language model predicts the tokens in each rationale. 
Perplexity is defined as the exponentiated average negative log-likelihood of a sequence. 
Note that perplexity is only for the generative baselines.

\paragraph{Token F1 (TF1)}
Following \citet{lakhotia-etal-2021-fid}, we compute the  Token-level F1 between ground-truth rationales and generated rationales.
TF1 measures the number of overlapping tokens between two rationales.
Following \citet{deyoung-etal-2020-eraser}, we use spaCy tokenizer\footnote{\url{https://spacy.io/}} to compute the F1 score.

\paragraph{Intersection over Union F1 (IOU F1)}
IOU F1, as used in \citet{deyoung-etal-2020-eraser}, computes the F1 on matched predictions.
IOU F1 first checks whether predicted rationales match ground-truth rationales by calculating the intersection of union (IOU). 
IOU is computed as the number of overlapping tokens divided by the union of tokens.
If IOU is larger than the threshold, the predicted explanation becomes a matched prediction.
In this work, we set the threshold as 0.5.

\begin{table*} [t]
\centering\
\resizebox{0.8\textwidth}{!}{
\def\arraystretch{0.9}
\begin{tabular}{l|cccc|cccc}
\toprule
\multirow{2}{*}{}
& \multicolumn{4}{c|}{\textit{\cdiffs}}& \multicolumn{4}{c}{\textit{\cdiffw}} \\
& AUROC & F1 & Precision & Recall &  
AUROC & F1 & Precision & Recall   \\\midrule
\textbf{\textit{finetuning}} & & & & & & & &\\
~RoBERTa-base & 0.6955 & 40.28 & 27.54 & 74.92 & 0.6303  & 19.70 & 13.17 & 39.08 \\
~RoBERTa-large & 0.7137 & 38.93 & 25.00 & \textbf{87.95} & 0.6646 & 19.75 & 11.79 & 60.92 \\ 
~RoBERTa (FEVER) & 0.7253 & 40.76 & 26.96 & 83.55 & 0.5981 & 15.35 & 8.54 & \textbf{75.63}\\ 
~RoBERTa (MNLI) & \textbf{0.7510} & 41.27 & 27.11 & 86.48 & 0.7005 & 24.46 & 17.11 & 42.86 \\ 
\midrule
\textbf{\textit{LoRA}} & & & & & & & & \\
~RoBERTa-large & 0.6620 & 36.57 & 24.11 & 75.73& 0.7282 & 24.17 & 18.05 & 36.55 \\
~DeBERTa-XXL & 0.7266 & 40.09 & 26.29 & 84.36 & \textbf{0.7289} & \textbf{26.70} & \textbf{19.06} & 44.54\\
\midrule
\textbf{\textit{zero-shot}} & & & & & & & & \\
~T0 & - & \textbf{41.77} & 34.09 & 53.91 & - & 13.37 & 8.10 & 38.24\\
~GPT-3 & - & 32.29 & \textbf{45.29} & 25.08 & - & 17.25 & 16.18 & 18.49\\
\bottomrule
\end{tabular}}
\vspace{-0.05cm}
\caption{Performance measured on test-doc split. 
The baselines are the same as those described in Section~\ref{sec:experiments_class}, but evaluated on a different test data.
}\vspace{-0.35cm}
\label{table:test-doc}
\end{table*}

\subsection{Results}

In Table~\ref{table:rationale}, we show the results of finetuned RoBERTa and T5 trained on rationale extraction.
Following \citet{deyoung-etal-2020-eraser}, we report extractive measures (TF1 and IOUF1), as the ground-truth rationales are extracted phrases from each claim. 
We further measure the generative score (perplexity) of output sequences for generative models. 
Note that we choose perplexity as the metric because rationales are in phrases rather than complete sentences.
Although \cdiff~contains multiple phrases as the ground-truth rationales, MRC models predict a single rationale for each claim pair. 
Since T5 is capable of generating multiple phrases at once, even smaller T5-base obtains better performance than RoBERTa-large.
T5-large consistently provides better results than T5-base regardless of whether the class labels are given or not. 
The injection of labels degrades the performance of T5-base, while the T5-large shows a slight improvement. 
The increasing number of parameters is also beneficial for incorporating additional label information.

\section{Extension: Document-level \cdiff}

\label{sec:application}
\miyoung{}
Suppose we want to compare the articles with opposing views on contentious issues. 
For instance, there are two articles about a topic, \textit{``Will Gas Prices Come Down Soon or Stay High?''}. 
One forecasts the \textit{increase} in gas prices, while the other supports the prices have \textit{already reached a peak}. 
A single stance label on the relation between the articles (i.e., whether one article supports or opposes the other) might not be enough to understand the complex relations of claims in the articles. 
\cdiff~can be applied to provide a granular-level comparison between two articles. 
Document-level \cdiff~enables to provide information about which arguments of the first article strengthen or weaken the views of the other. 
We provide a live demo for document-level extension with RoBERTa-large model.\footnote{\url{https://www.claimdiff.com}}
As an example, the demo result of the above topic is shown in Appendix~\ref{sec:demo}.

In order for the real-world document-level comparison scenario, we evaluate our baseline models on the test-doc dataset, which follows the real-world label distribution. Specifically, the test-doc dataset includes all non-overlapping sentence pairs, which were originally filtered out for the test dataset construction as described in Section~\ref{sec3:construction}. 
The results are shown in Table \ref{table:test-doc}.
Note that we do not provide human performance on the test-doc, as obtaining human annotation over the whole article is costly.
Aligning with previous observations, RoBERTa (MNLI) and LoRA with the DeBERTa-XXL achieves the best AUROC on document-level \cdiffs~and \cdiffw, respectively.
However, unlike previous results, T0 achieves 
the second-best F1 on test-doc \cdiffs.
One possible reason in that finetuned models have a high proportion of false positives in the full document setting due to distribution shift, whereas the zero-shot model seems to be more robust to it.

Although the fine-grained comparison is helpful for understanding contentious issues, looking over the whole article pair is costly.  
Future work includes providing summarized statistics from fine-grained comparisons.
For example, the ratio of strengthening / weakening pairs can represent how much the two articles oppose each other. 
We can further extend \cdiff~to compare more than two articles with summarized results, and even compare between different presses. 

\section{Conclusion}
\label{sec:conclusion}

This paper presents \cdiff~, a new benchmark dataset of 2.9k annotation to compare claims in news articles on contentious issues. 
Unlike the previous fact verification, \cdiff~focues on \textit{comparing} the nuance between claim pairs from trusted sources, whether one claim \textit{strengthens} or \textit{weakens} the other.
We experiment with pre-trained language models in finetuning, parameter-efficient finetuning, and zero-shot approaches. 
The results show a significant room for improvement with over 19\% absolute gap between human and model performance. 
We further suggest document-level \cdiff~ as a real-world application and show its potential by presenting the baseline performance on the test-doc dataset that follows the real-world distribution.
We hope this initial study could pave the way for providing an analysis tool for article readers to obtain an unbiased understanding of contentious issues.

\section*{Limitations}
First, most articles are crawled from the US and UK presses. This means the crawled data is English-only and regionally biased, limiting the scope and the diversity of issues. Extending our work to other languages and more regionally-diverse presses will be helpful for reducing such bias in our dataset.

Second, we suspect that there will be a non-trivial annotation bias in our dataset. We are concerned with the fact that all of our in-house annotators share the same cultural background and similar personal interest (given that the annotators volunteered to partcipate in this turking task). Furthermore, given that ClaimDiff-W is aiming to catch the subtle differences in the nuances of these professional news articles, it is very challenging for different annotators to have a common view, especially compared to ClaimDiff-S (which also explains why ClaimDiff-W human performance is much lower than that of ClaimDiff-S). 

Third, since ClaimDiff is a sentence-level comparison task, it currently does not give information about the surrounding context of each sentence. This means inter-sentence dependency such as coreference often cannot be resolved. One way to work around this is to give an access to the full articles for each claim pair, but we have refrained from it in this work for simplicity (though we believe it will be interesting to see if the performance can be improved with such access).

Fourth, the size of ClaimDiff is relatively small compared to other fact verficiation datasets. This is mainly because its annotation process is quite challenging and requires a substantial amount of time. Future work includes expanding the size of ClaimDiff when additional budget is available.

\section*{Acknowledgements}
We thank Yongrae Jo, Joel Jang, Hyunji Lee, Hanseok Oh, Soyoung Yoon, Hwisang Jeon and Gangwoo Kim for the useful discussion and feedback on the paper.
This work was partly supported by KAIST-NAVER Hypercreative AI Center (80\%) and Institute of Information \& communications Technology Planning \& Evaluation (IITP) grant funded by the Korea government (MSIT) (No.2021-0-02068, Artificial Intelligence Innovation Hub, 20\%).


\bibliography{custom}
\bibliographystyle{acl_natbib}

\clearpage
\newpage
\appendix
\label{sec:appendix}

\section{Data Construction Process}\label{appendix:data_collect}
\label{appendix:data_collect}

This section describes the preprocessing step and the data construction process by human annotators.
We construct \cdiff~by 2 steps:
(i) filtering non-overlapping claim pairs and (ii) annotating the pairs.

\paragraph{Removing Identifying Information} 
We first preprocess the collected articles to remove identifying information, such as reporters' contact information. 
We remove personal information in a two-step procedure. 
First, we use automatic ways—regular expressions and pre-trained language models—to classify whether a sentence contains personal information. 
Then, after the automatic removal step, we manually inspect each sentence again and remove the sentence if it contains personal information.

\begin{figure*}[t]
    \centering
    \includegraphics[width=0.95\textwidth]{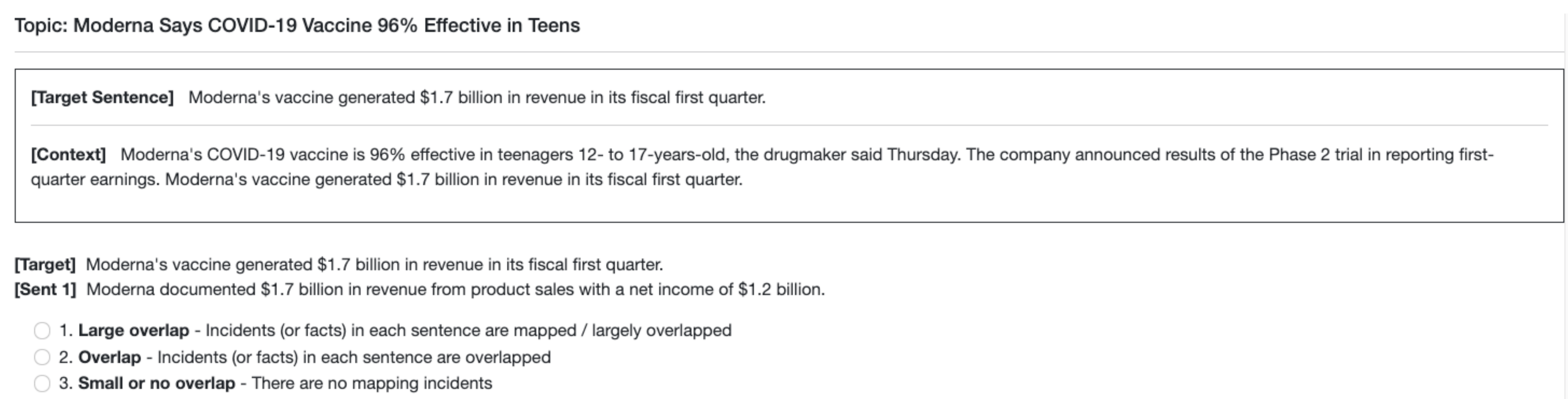}
    \caption{Interface for filtering task.}\vspace{-0.2cm}
    \label{fig:appendix_amt}
\end{figure*}

\paragraph{Data Filtering Process}we use MTurk\footnote{\url{https://www.mturk.com}} for filtering large amount of non-overlapping claims. 
Each worker is asked to solve Human Intelligent Tasks (HITs), which consist of 6 multiple-choice questions. 
Each HITs is composed of 1 quiz question to manage workers and 5 claim pairs extracted from news articles. 
The interface for a single question is presented in Figure~\ref{fig:appendix_amt}.
The reward for a single HIT is \$0.18.
We collect the responses from 3 different workers for a single example. 
If more than two workers choose the 'overlap' or 'large overlap', the pair are then considered as the 'overlapping' pair. 
If more than two workers choose 'small or no overlap', then the pair is considered as 'non-overlapping'.
We process the annotation step for only the 'overlapping' claim pairs.

\begin{figure*}[t]
    \centering
    \includegraphics[width=0.95\textwidth]{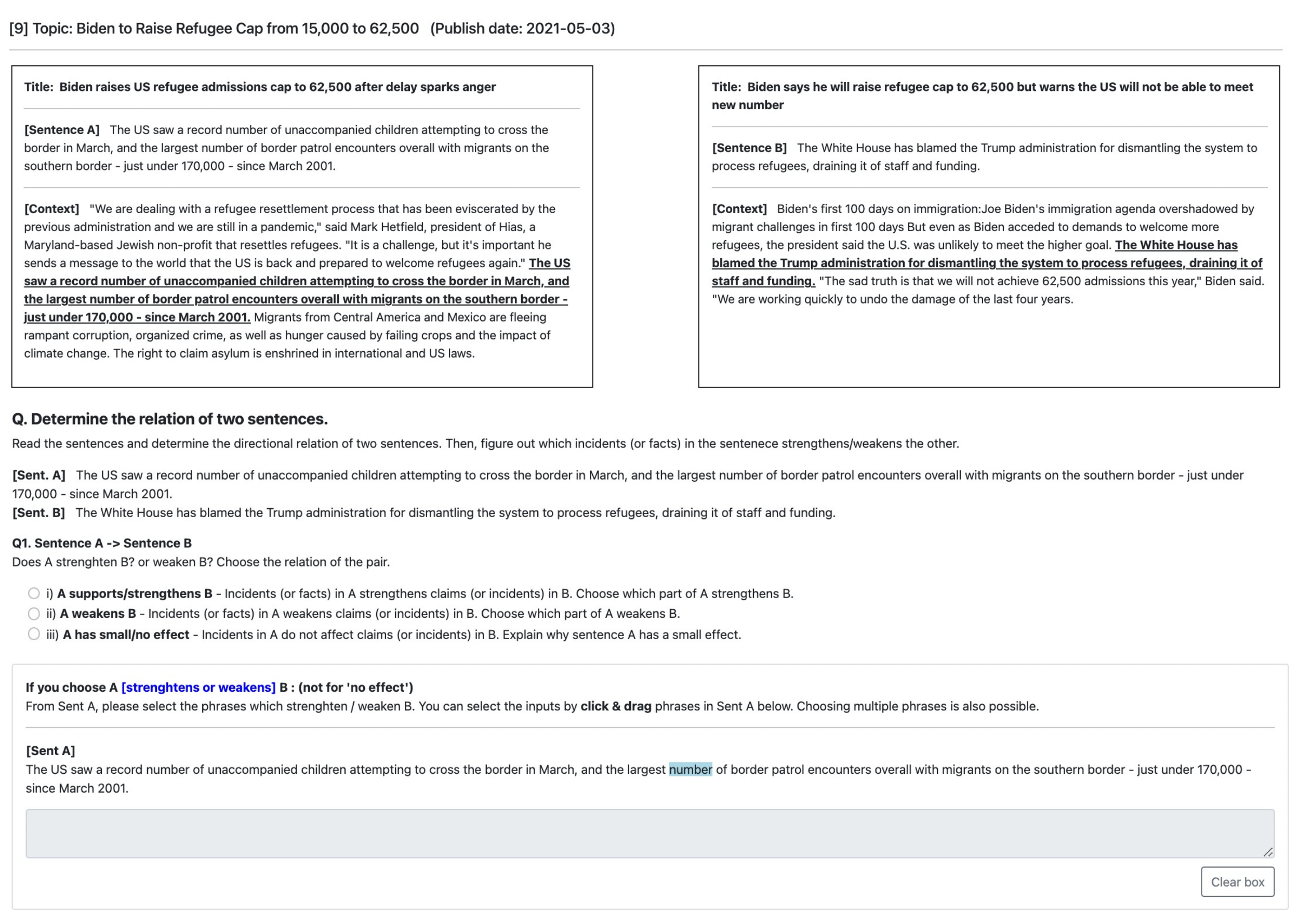}
    \caption{Screenshot of annotation task.}\vspace{-0.2cm}
    \label{fig:appendix_annot}
\end{figure*}

\paragraph{Data Annotation Process} For the second annotation step, we separately hire 15 in-house expert annotators.
We held two training sessions for in-house experts; one for providing guidelines and the other for solving example tasks. 
Each expert should pass the final quiz (15 out of 16 questions) after training sessions to start the main tasks. 
The interface for annotation is shown in Figure~\ref{fig:appendix_annot}.
Annotators are asked to choose the directional relation of a given pair and select the rationale that supports the relation. 
In the data construction process, we provide additional context information for a better understanding of the sentence.
We offer \$0.25 for a single example.\footnote{We provide at least \$7.5 per hour even if annotators submit less than 30 responses.}

\begin{table}
\centering\
\resizebox{0.43\textwidth}{!}{
\def\arraystretch{0.9}
\begin{tabular}{l|ccc}
\toprule
\multirow{2}{*}{}
\textbf{Model} & \textbf{F1} & \textbf{Precision} & \textbf{Recall} \\\midrule
\multicolumn{4}{c}{\textit{\cdiffs}} \\\midrule
RoBERTa-base& 81.87 & 76.55 & 87.99  \\
RoBERTa-large& 84.70 & 76.17 & 95.38 \\
RoBERTa (FEVER) & 80.38 & 68.58 & 97.08 \\
RoBERTa (MNLI)& 83.12& 78.26 & 88.62 \\
LoRA (RoBERTa)& 82.91 & 77.19 & 89.54 \\
LoRA (DeBERTa) & 82.60 & 75.68 & 90.91
\\\midrule
\multicolumn{4}{c}{\textit{\cdiffw}} \\\midrule
RoBERTa-base& 43.24& 34.78 & 57.14  \\
RoBERTa-large& 38.37 & 25.38 & 78.57 \\
RoBERTa (FEVER) & 18.29 & 10.39 & 76.19 \\
RoBERTa (MNLI)& 59.02& 50.70 & 70.59 \\
LoRA (RoBERTa)& 43.18& 41.30 & 45.24 \\
LoRA (DeBERTa) &56.18& 53.19 & 59.52 \\
\bottomrule
\end{tabular}}
\caption{Validation performance of finetuning and parameter-efficient finetuning baselines.
}
\label{table:validation}
\end{table}

\section{Implementation Details}\label{appendix:imp_detail}
For all experiments, we use PyTorch \cite{NEURIPS2019_bdbca288} and Transformers \cite{wolf-etal-2020-transformers}.
Most of the experiments are conducted on 8 V100 GPUs except T5 for rationale extraction. The validation performance of each model is presented in Table~\ref{table:validation}.

\begin{table} [t]
\centering
\resizebox{0.43\textwidth}{!}{
\def\arraystretch{0.9}
\begin{tabular}{l|c}
\toprule
\textbf{Hyperarameter} & \textbf{Search space} \\\midrule
Learning rate & \{5e-4, 1e-4, 5e-5, 1e-5, 5e-6, 1e-6\}\\
Warmup steps & \{0, 50, 100, 150, 200\} \\
Weight decay & \{off, 1e-5\}\\
\bottomrule
\end{tabular}}
\caption{Search space for hyperparameters of finetuned RoBERTa.}\vspace{-0.4cm}
\label{table:search_space}
\end{table}

\subsection{RoBERTa (MNLI) /  RoBERTa (FEVER)}
RoBERTa (MNLI) and RoBERTa (FEVER) are finetuned RoBERTa-large models finetuned on MNLI and FEVER, respectively.
We load the trained checkpoints for the MNLI model,\footnote{\url{https://huggingface.co/roberta-large-mnli}} while me manually finetune RoBERTa-large\footnote{\url{https://huggingface.co/roberta-large}} for FEVER.  
Following ~\citet{nie2019combining}, we convert FEVER into an NLI-style task, predicting only the labels among the given query and context. 
We train RoBERTa-large on NLI-style FEVER during 10 epochs with batch size of 32 and 100 warmup steps. 
The model is optimized by Adam\citep{DBLP:journals/corr/KingmaB14} optimizer with learning rate of $5e^{-5}$.

\subsection{Relation Classification}

\paragraph{Fientuned RoBERTa} 
For finetuning experiments, we train models during 10 epochs with a batch size of 32 and 200 warmup steps. 
We find the best hyperparameters for each model using the results of 3-fold cross-validation. 
We choose the best-working checkpoints and thresholds based on the validation F1 score. 
The search space for each hyperparameter is presented in Table~\ref{table:search_space}.
We use Adam optimizer for training. 
For \cdiffs, we use learning rates of $5e^{-5}$ for RoBERTa-base and $5e^{-6}$ for others.
\cdiffw~models are trained with learning rate of $1e^{-5}$ for RoBERTa (MNLI), and $5e^{-6}$ for other models.

\paragraph{LoRA} 
We train RoBERTa-large and DeBERTa-XXL\footnote{\url{https://huggingface.co/microsoft/deberta-v2-xxlarge}}  with LoRA during 20 epochs with a batch size of 32, the learning rate of $1e^{-4}$, and 0.01 weight decay. 
We use the LoRA implementation released by the authors\footnote{\url{https://github.com/microsoft/LoRA}}. 
We use a linear scheduler for the learning rate schedule with a 0.1 warmup ratio.
We set the rank as 8 with LoRA $\alpha$ of 16 for RoBERTa-base. 
For DeBERTa-XXL, rank and $\alpha$ are set to be 16 and 32.

\paragraph{T0}

As the proposed tasks are binary classification tasks, T0 takes the input prompts and generates answers between ('yes', 'no') as prediction labels.
We use pre-trained weights of T0\footnote{\url{https://huggingface.co/bigscience/T0}} for zero-shot prediction. 
The input prompts for \cdiffs~and \cdiffw~are as follow: 

\noindent(i) \cdiffs
\begin{verbatim}
Claim A: {claim_a} \n\n
Claim B: {claim_b} \n\n
Does Claim A support Claim B? yes or no?
\end{verbatim}

\noindent(ii) \cdiffw
\begin{verbatim}
Claim A: {claim_a} \n\n
Claim B: {claim_b} \n\n
Does Claim A weaken Claim B? yes or no?
\end{verbatim}

\paragraph{GPT-3}
Similar to T0, we consider predictions of GPT-3 as correct if GPT-3 generates 'Yes' (when the label is 1) or 'No' (when the label is 0) in the outputs.
We use \textit{text-davinci-003} of the GPT-3 family in this work. 
The input prompts for \cdiffs~and \cdiffw~are as follow: 

\noindent(i) \cdiffs
\begin{verbatim}
Does A support B?: \n\n
A: {claim_a} \n\n B: {claim_b} \n\n
\end{verbatim}

\noindent(ii) \cdiffw
\begin{verbatim}
Does A weaken B?: \n\n
A: {claim_a} \n\n B: {claim_b} \n\n
\end{verbatim}

\subsection{Rationale Extraction}

\paragraph{MRC Models}
We train RoBERTa-base and RoBERTa-large with additional answer prediction layers during 3 epochs.
Models are trained with Adam optimizer with a learning rate of $5e^{-5}$ and batch size of 32.
We choose the best checkpoints based on validation TF1.

\paragraph{T5}
We finetune T5-base\footnote{\url{https://huggingface.co/t5-base}} and T5-large\footnote{\url{https://huggingface.co/t5-large}} to directly generate a list of rationales. 
More formally, given claim pairs $(c_1, c_2)$,  we optimize models to obtain the list of rationale phrases. 
The model takes input as "explain claimdiff claim1: $c_1$ claim2: $c_2$", and is trained to generate the target sequence represented as "explanation: \{rationale1\} explanation: \{rationale2\} ···".
For \textit{class label} models, we additionally append "relation: $r$" to the input text as class information, where $r$ is either one of \textit{strengthen} or \textit{weaken}.

We use T5 with a maximum input sequence length of 512 and a batch size of 8. 
All experiments are conducted on 4 Tesla M60 GPUs using ZeRO \citep{rajbhandari2019zero} stage-3 provided in DeepSpeed  \citep{rasley2020deepspeed} to reduce GPU memory usage. 
We train all models using the Adam optimizer with a constant learning rate of $1e^{-4}$.
To obtain rationales, we perform beam search decoding using a beam size of 2.

\section{\cdiff~Statisics}\label{appendix:stat}
Table~\ref{table:appendix_stat} shows top-15 presses and tags with their occurrence. 
Tags in \cdiff~have long-tailed distribution, indicating \cdiff~do not concentrate of specific topic. 

\section{\cdiff~Examples}\label{appendix:example}
The examples positive and negative examples of \cdiff~are presented in Table~\ref{table:appendix_example}.
Note that the same pair can have different labels for \cdiffs~and \cdiffw.

\section{Dataset Analysis}\label{appendix:data_anal}
\begin{figure*} [t]
    \centering
     \subfloat[Subjectivity Analysis]
     {\includegraphics[width=0.38\textwidth]{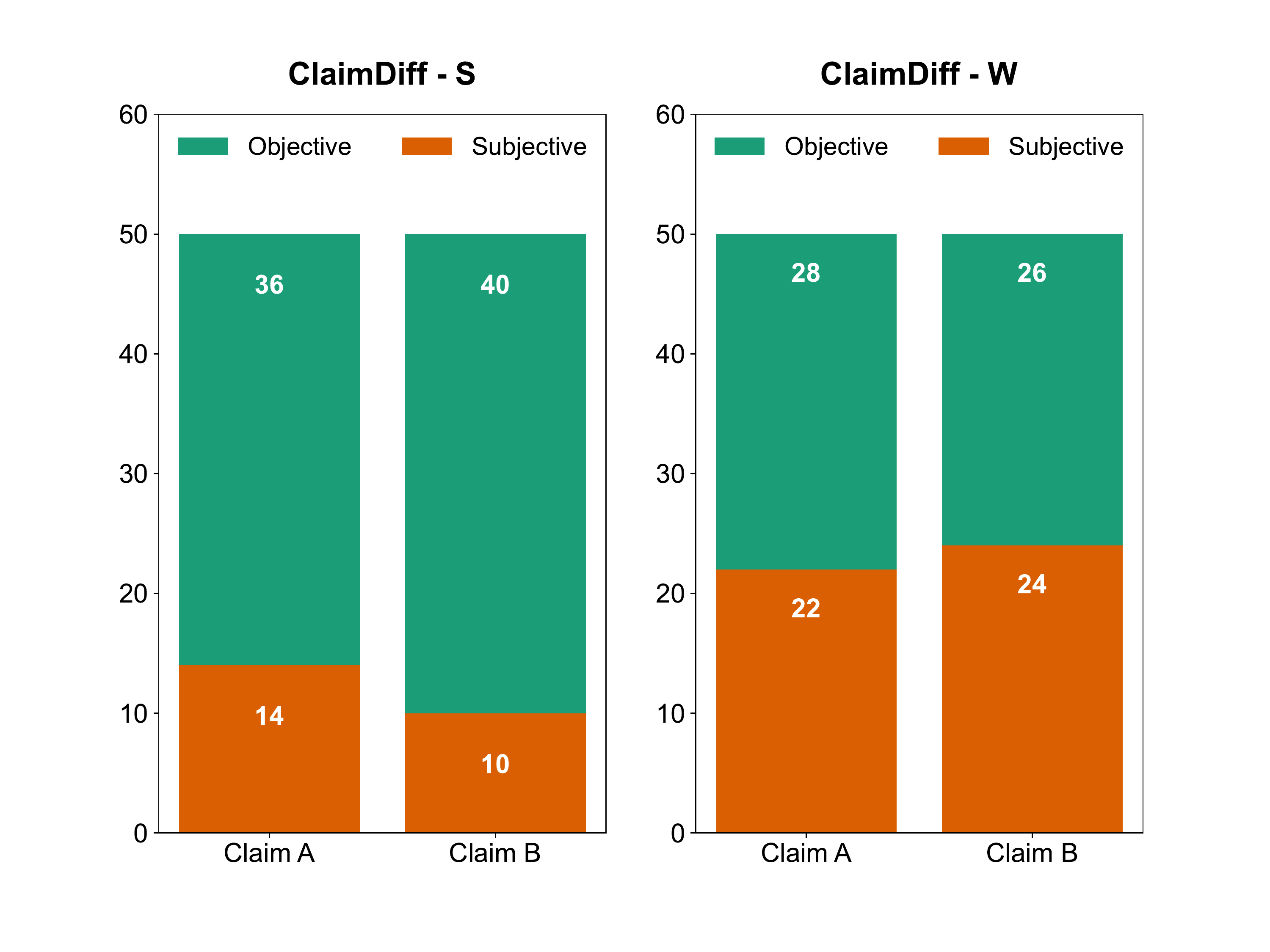}}
     \subfloat[MNLI Model Predictions]
     {\includegraphics[width=0.31\textwidth]{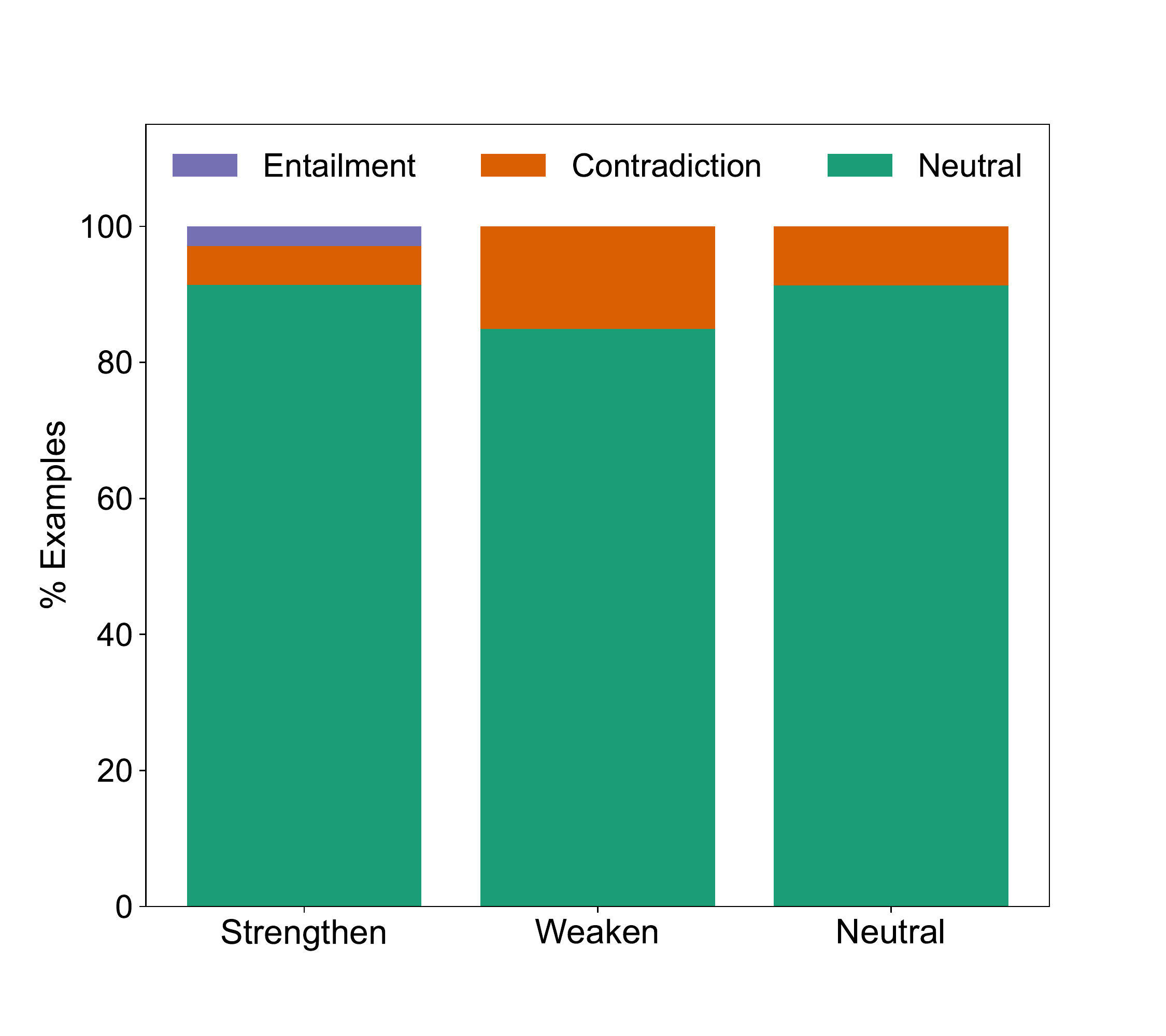}}
     \subfloat[FEVER Model Predictions]
     {\includegraphics[width=0.31\textwidth]{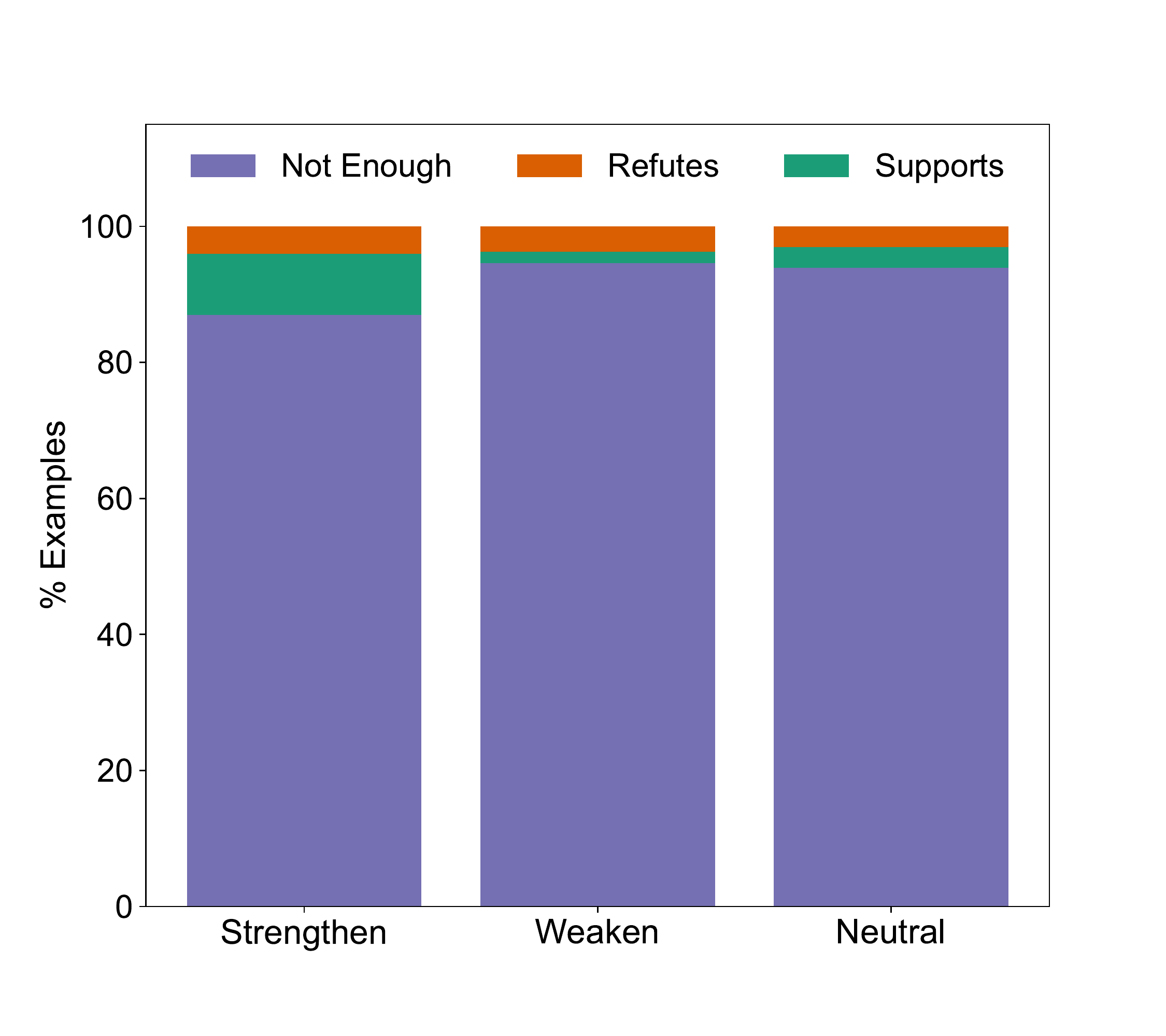}}
     \vspace{-0.05cm}
    \caption{ Analysis over claim pairs in \cdiff. (a) Subjectivity results for strengthening and weakening pairs with positive labels. (b) Prediction results of Roberta-large trained on MNLI dataset. (c) Prediction results of Roberta-large trained on FEVER.}\vspace{-0.35cm}
    
    \label{fig:fig3}
\end{figure*}

Figure~\ref{fig:fig3} provides more detailed results of Section~\ref{sec3:analysis}.
For MNLI model, we gave the former claim as 'premise' and the later claim as 'hypothesis. 
For FEVER model predictions, the second claim are given as 'claim' and the former as 'evidence'.


\section{Error Analysis}\label{appendix:error_analysis}
We randomly sample 25 false negatives of RoBERTa-large predictions from \cdiffs~and \cdiffw, respectively. 
Table~\ref{table:appendix_error_s} and Table~\ref{table:appendix_error_w} show each error category and its corresponding example in false negative errors.

\section{Live Demo Result}
\label{sec:demo}
\begin{figure*} [t]
    \centering
     \includegraphics[width=0.7\textwidth]{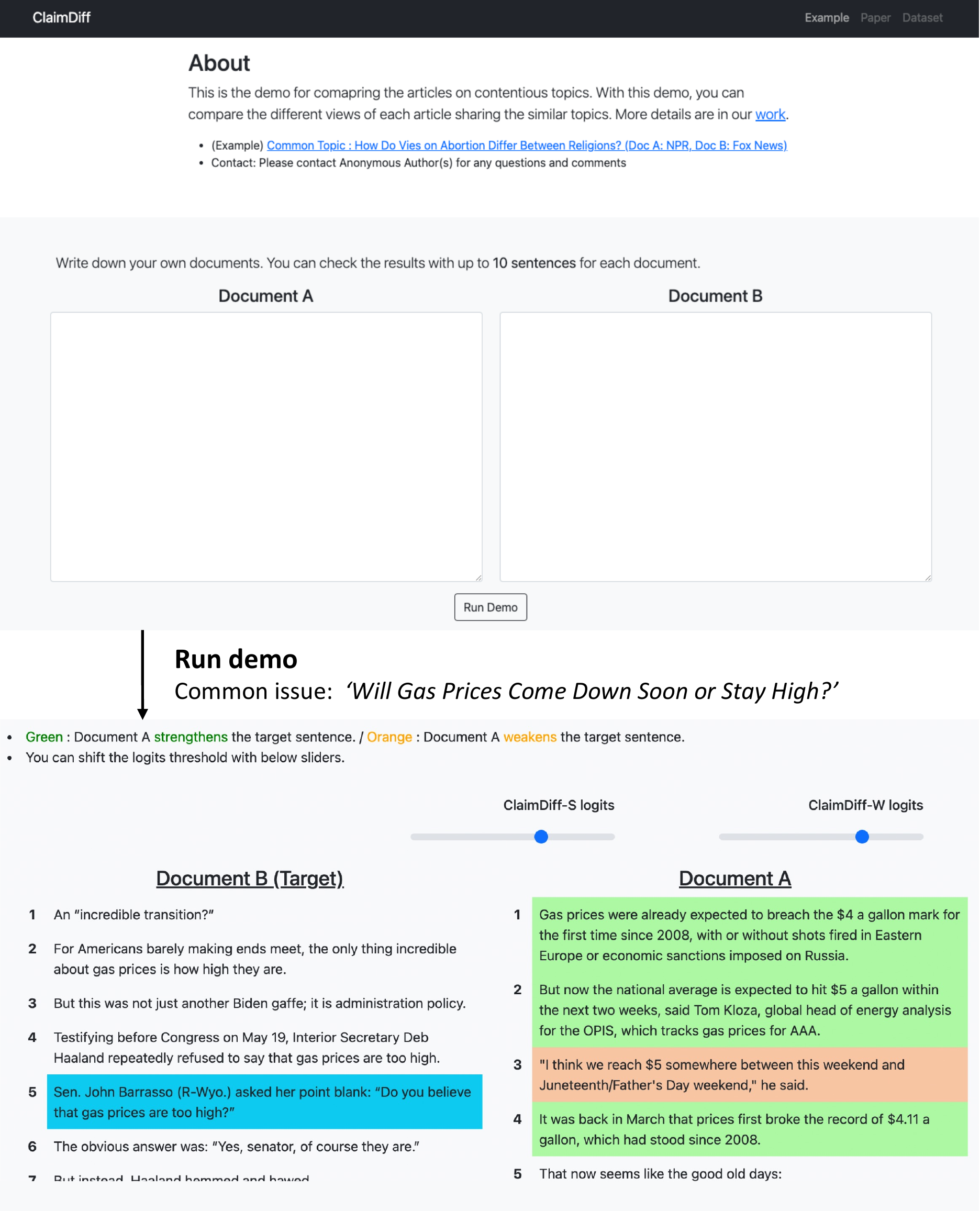}
    \caption{Demo results on the two articles about a topic, \textit{'Will Gas Prices Come Down Soon or Stay High?}'. 
    Sentences in green represent the claims that strengthen the 5-th sentence of document B.
    Sentence in orange indicates the claim that weakens the sentence of document B.}\vspace{-0.3cm}
    \label{fig:appendix_demo}
\end{figure*}

Figure~\ref{fig:appendix_demo} shows the screenshot of the demo and running results.

\newpage

\begin{table*} [t]
\centering\
\resizebox{0.9\textwidth}{!}{
\def\arraystretch{0.9}
\begin{tabularx}{\textwidth}{c|X}
\toprule
\textbf{Press} & 
Fox News (Online News) (32), CNN (Online News) (27), Washington Times (25), Politico (17), Associated Press (15), USA TODAY (12), New York Post (News) (11), NBC News (Online) (11), Reuters (11), The Hill (9), Vox (8), NPR (Online News) (8), The Guardian (7), National Review (6), New York Times (News) (6) \\\midrule
\textbf{Tag} & 
Elections (14), DonaldTrump (13), coronavirus (11), USSenate (9), Immigration (9), PresidentialElections (8), Technology (7), SupremeCourt (7), World (6), ViolenceinAmerica (5), MediaBias (5), WhiteHouse (5), Russia (5), MiddleEast (5), Business (5) \\
\bottomrule
\end{tabularx}}
\caption{Top-15 presses and tags included in \cdiff.
The numbers indicate the occurrence of each press or tag.}
\label{table:appendix_stat}
\end{table*}

\begin{table*} [t]
\centering\
\resizebox{\textwidth}{!}{
\def\arraystretch{0.9}
\begin{tabularx}{\textwidth}{c|X}
\toprule
\textbf{Label} & \textbf{Example} \\\midrule
\multicolumn{2}{c}{\textcolor{teal}{\textit{\cdiffs}}}\\\midrule
\multirow{4}{*}{\textbf{1}} &
\small\textbf{Issue}: First-of-its-kind California Program Offers Virus Aid to People in the Country Illegally \\
&\small\textbf{Claim A}: Legal complaints lodged to try to stop the distribution of funds to illegal aliens were blocked, one by the California Supreme Court on May 6 and one by the Los Angeles Superior Court on May 5. \\
&\small\textbf{Claim B}: Applicants are eligible for the money if they demonstrate they are unauthorized, jobless as a result of the pandemic, and do not qualify unemployment programs or stimulus checks.\\
&\small\textbf{Rationale}: [Legal complaints lodged to try to stop the distribution of funds to illegal aliens were blocked,] \\\midrule
\multirow{4}{*}{\textbf{1}} &
\small\textbf{Issue}: Pfizer Says its COVID-19 Vaccine is Safe, Effective for Kids Ages 5-11  \\
&\small\textbf{Claim A}: Coronavirus infections have risen "exponentially" among children across the United States, and now account for nearly 29\% of all cases reported nationwide, the American Academy of Pediatrics reported last week. \\
&\small\textbf{Claim B}: "Since July, pediatric cases of COVID-19 have risen by about 240 percent in the U.S. - underscoring the public health need for vaccination," Pfizer's CEO Albert Bourla said in a statement. \\
&\small\textbf{Rationale}:[Coronavirus infections have risen "exponentially" among children across the United States,] \\\midrule
\multirow{4}{*}{\textbf{0}} &
\small\textbf{Issue}: Hack Cuts Off Nearly 20\% of US Meat Production  \\
&\small\textbf{Claim A}: Any further impact on consumers will depend on how long JBS plants remain closed, analysts said.\\
&\small\textbf{Claim B}: The Colonial Pipeline, which provides 45\% of the gas used in East Coast states, was hacked and temporarily shut down by East European hacker group DarkSide. \\
&\small\textbf{Rationale}: [] \\\midrule
\multirow{4}{*}{\textbf{0}} &
\small\textbf{Issue}: FDA Commissioner Acknowledges Misrepresenting Convalescent Plasma Data \\
&\small\textbf{Claim A}: The FDA made the decision based on data the Mayo Clinic collected from hospitals around the country that were using plasma on patients in wildly varying ways and there was no comparison group of untreated patients, meaning no conclusions can be drawn about overall survival. \\
&\small\textbf{Claim B}: Speaking at that press conference, Trump claimed that blood plasma treatment had cut COVID-19 mortality by 35\%. \\
&\small\textbf{Rationale}: [] \\\midrule
\multicolumn{2}{c}{\textcolor{orange}{\textit{\cdiffw}}}\\\midrule
\multirow{4}{*}{\textbf{1}} &
\small\textbf{Issue}: FDA Commissioner Acknowledges Misrepresenting Convalescent Plasma Data \\
&\small\textbf{Claim A}: The FDA made the decision based on data the Mayo Clinic collected from hospitals around the country that were using plasma on patients in wildly varying ways and there was no comparison group of untreated patients, meaning no conclusions can be drawn about overall survival. \\
&\small\textbf{Claim B}: Speaking at that press conference, Trump claimed that blood plasma treatment had cut COVID-19 mortality by 35\%. \\
&\small\textbf{Rationale}: [using plasma on patients in wildly varying, there was no comparison group of untreated patients, no conclusions can be drawn about overall survival.] \\\midrule
\multirow{4}{*}{\textbf{1}} &
\small\textbf{Issue}: Facebook Changes Trending News\\
&\small\textbf{Claim A}: In a poll conducted by the media and data analysis site Morning Consult, only 48 percent of respondents said they had heard about the bias allegations against Facebook. \\
&\small\textbf{Claim B}: But the company also runs a "Trending Topics" section that promotes some stories, and that's where the bias charges focused. \\
&\small\textbf{Rationale}: [only 48 percent of respondents said they had heard about the bias allegations] \\\midrule
\multirow{4}{*}{\textbf{0}} &
\small\textbf{Issue}: Hack Cuts Off Nearly 20\% of US Meat Production  \\
&\small\textbf{Claim A}: Any further impact on consumers will depend on how long JBS plants remain closed, analysts said.\\
&\small\textbf{Claim B}: The Colonial Pipeline, which provides 45\% of the gas used in East Coast states, was hacked and temporarily shut down by East European hacker group DarkSide. \\
&\small\textbf{Rationale}: [] \\\midrule
\multirow{4}{*}{\textbf{0}} &
\small\textbf{Issue}: First-of-its-kind California Program Offers Virus Aid to People in the Country Illegally \\
&\small\textbf{Claim A}: Legal complaints lodged to try to stop the distribution of funds to illegal aliens were blocked, one by the California Supreme Court on May 6 and one by the Los Angeles Superior Court on May 5. \\
&\small\textbf{Claim B}: Applicants are eligible for the money if they demonstrate they are unauthorized, jobless as a result of the pandemic, and do not qualify unemployment programs or stimulus checks.\\
&\small\textbf{Rationale}: [] \\
\bottomrule
\end{tabularx}}
\caption{Examples of \cdiffs~and \cdiffw.}
\label{table:appendix_example}
\end{table*}

\begin{table*} [t]
\centering\
\resizebox{\textwidth}{!}{
\def\arraystretch{0.9}
\begin{tabularx}{\textwidth}{X}
\toprule
\multicolumn{1}{c}{\textcolor{teal}{\textit{\cdiffs}}}\\\midrule
\small\textbf{\textcolor{teal}{(7 / 25)}\underline{ [Category 1] Entailment or Objective Example}} \\
\small\textbf{Issue}: CDC Shortens School Distancing Guidelines to 3 Feet with Masks \\
\small\textbf{Claim A}: Dr. Lawrence Kleinman, ..., said 3 feet is "probably safe" if schools are doing everything right - if everyone is wearing masks correctly at all times and washing their hands, and if ventilation is good. \\
\small\textbf{Claim B}: In Utah , a study found that 86\% of students wore masks in elementary school classrooms and very few passed the virus to others. \\\midrule
\small\textbf{\textcolor{teal}{(4 / 25)}\underline{ [Category 2] Cuase-and-Effect}}  \\
\small\textbf{Issue}: Congress Estimates Social Security to Run Out by 2031 \\
\small\textbf{Claim A}: Since no one is suggesting raising taxes to make up the lost revenue from Social Security, that additional \$1 trillion would have more than doubled the fiscal year 2019 deficit. \\
\small\textbf{Claim B}: But, ... removing the FICA tax as a funding source for Social Security, ... requires an increase in tax revenue of some fashion or another. \\\midrule
\small\textbf{\textcolor{teal}{(5 / 25)}\underline{ [Category 3] Sharing the same nuances}}  \\
\small\textbf{Issue}: FDA Commissioner Acknowledges Misrepresenting Convalescent Plasma Data \\
\small\textbf{Claim A}: Trump hailed the decision as a historic breakthrough even though the treatment's value has not been established. \\
\small\textbf{Claim B}: The president also claimed that plasma "had an incredible rate of success" for treating COVID-19 patients, despite the fact that his own scientists and the FDA itself had expressed more reserved assessments. \\\midrule
\small\textbf{\textcolor{teal}{(3 / 25)}\underline{ [Category 4] Lack of Context Information}}  \\
\small\textbf{Issue}: How the 4/20 Holiday is Celebrated During Coronavirus Pandemic\\
\small\textbf{Claim A}: If you've been feeling anxiety over current events, it might be time to browse a few soothing CBD products - especially in honor of this week's cheeky 4/20 holiday, celebrated by cannabis lovers around the globe.\\
\small\textbf{Claim B}: Part of that may have been in preparation for 4/20 celebrations.\\\midrule
\small\textbf{\textcolor{teal}{(2 / 25)}\underline{ [Category 5] Supporting interview}}  \\
\small\textbf{Issue}: Jeff Sessions Hits Back At Trump \\
\small\textbf{Claim A}: During a Thursday morning interview on "Fox \& Friends," Mr. Trump renewed his criticism of Mr. Sessions, accusing him of allowing the Justice Department to undermine his administration. \\
\small\textbf{Claim B}: "No, the truly unique thing here is that Sessions decided to actually speak up in his own defense." \\
\bottomrule
\end{tabularx}}
\caption{Cateogries and correspoding examples of false negatives in \cdiffs. The number indicates how many examples fall into the category over 25 examples.}
\label{table:appendix_error_s}
\end{table*}

\begin{table*} [t]
\centering\
\resizebox{\textwidth}{!}{
\def\arraystretch{0.9}
\begin{tabularx}{\textwidth}{X}
\toprule
\multicolumn{1}{c}{\textcolor{orange}{\textit{\cdiffw}}}\\\midrule
\small\textbf{\textcolor{orange}{(15 / 25)}\underline{ [Category 1] Contradiction / Conflicts - (Type 1) Contradiction}}  \\
\small\textbf{Issue}: Election Systems Hacked by Russians \\
\small\textbf{Claim A}: In this instance, the username and password information posted would only give individuals access to a localized, county version of the voting registration system, and not the entire state-wide system. \\
\small\textbf{Claim B}: Hackers based in Russia were behind two recent attempts to breach state voter registration databases, fueling concerns the Russian government may be trying to interfere in the U.S. presidential election, U.S. intelligence officials tell NBC News. \\\midrule
\small\textbf{\textcolor{orange}{(15 / 25)}\underline{ [Category 1] Contradiction / Conflicts - (Type 2) Conflicting Arguments}}  \\
\small\textbf{Issue}: FDA Commissioner Acknowledges Misrepresenting Convalescent Plasma Data \\
\small\textbf{Claim A}: Though scientists and medical experts are in agreement that the emergency authorization would likely make it easier for certain hospitals and clinics to access plasma, a promising treatment strategy which uses antibodies of recovered patients, many expressed alarm Sunday over Trump's rhetoric. \\
\small\textbf{Claim B}: Hahn had echoed Trump in saying that 35 more people out of 100 would survive the coronavirus if they were treated with the plasma. \\\midrule
\small\textbf{\textcolor{orange}{(3 / 25)}\underline{ [Category 2] Lack of Context Information}}  \\
\small\textbf{Issue}: Negotiating the Fiscal Cliff \\
\small\textbf{Claim A}: Obama expressed optimism as he took \underline{his case on the road here Friday}, saying Democrats and Republicans "can and will work together." \\
\small\textbf{Claim B}: The remarks came a day after the Obama administration unveiled details of a comprehensive package, widely rejected by Republicans, to avert the fiscal cliff. \\\midrule
\small\textbf{\textcolor{orange}{(2 / 25)}\underline{ [Category 3] Opposing nuances}}  \\
\small\textbf{Issue}: CDC Issues Guidance for Fully Vaccinated Individuals \\
\small\textbf{Claim A}: The guidance was "welcome news to a nation that is understandably tired of the pandemic and longs to safely resume normal activities," said Dr. Richard Besser, president and CEO of the Robert Wood Johnson Foundation and a former acting director of the CDC. \\
\small\textbf{Claim B}: She stressed that everyone should continue to avoid nonessential trips, regardless of vaccination status. \\
\bottomrule
\end{tabularx}}
\caption{Cateogries and correspoding examples of false negatives in \cdiffw. The number indicates how many examples fall into the category over 25 examples.
Note that there are diverse patterns in contradiction / conflicts category, which makes \cdiffw~more challenging.
We present two types contradiction / conflicts as examples.}
\label{table:appendix_error_w}
\end{table*}

\end{document}